\documentclass[10pt,twocolumn,letterpaper]{article}

\usepackage{cvpr}              

\makeatletter
\@namedef{ver@everyshi.sty}{}
\makeatother
\usepackage{tikz}

\usepackage{graphicx}
\usepackage{amsmath}
\usepackage{amssymb}
\usepackage{booktabs}
\usepackage{pgfplots}
\usepackage{float}

\newcommand{\customwidth}{\textwidth}

\usepackage[pagebackref,breaklinks,colorlinks]{hyperref}

\usepackage[capitalize]{cleveref}
\crefname{section}{Sec.}{Secs.}
\Crefname{section}{Section}{Sections}
\Crefname{table}{Table}{Tables}
\crefname{table}{Tab.}{Tabs.}
\Crefname{figure}{Figure}{Figures}
\crefname{figure}{Fig.}{Figs.}

\DeclareUnicodeCharacter{200E}{}

\def\cvprPaperID{*****} 
\def\confName{CVPR}
\def\confYear{2023}

\begin{document}

\title{LDM3D: Latent Diffusion Model for 3D}


\author{Gabriela Ben Melech Stan\\
Intel Labs\\
{\tt\small gabriela.ben.melech.stan@intel.com}
\and
Diana Wofk\\
Intel Labs\\
{\tt\small diana.wofk@intel.com}
\and
Scottie Fox\\
Blockade Labs\\
{\tt\small scottie@blockadelabs.com}
\and
Alex Redden\\
Blockade Labs\\
{\tt\small alexander.h.redden@gmail.com}
\and
Will Saxton\\
Blockade Labs\\
{\tt\small imagearts360@gmail.com}
\and
Jean Yu\\
Intel\\
{\tt\small jean1.yu@intel.com}
\and
Estelle Aflalo\\
Intel Labs\\
{\tt\small estelle.aflalo@intel.com}
\and
Shao-Yen Tseng\\
Intel Labs\\
{\tt\small shao-yen.tseng@intel.com}
\and
Fabio Nonato\\
Intel\\
{\tt\small fabio.nonato.de.paula@intel.com}
\and
Matthias M\"uller\\
Intel Labs\\
{\tt\small matthias.mueller@intel.com}
\and
Vasudev Lal\\
Intel Labs\\
{\tt\small vasudev.lal@intel.com}
}

\maketitle

\begin{abstract}
This research paper proposes a Latent Diffusion Model for 3D (LDM3D) that generates both image and depth map data from a given text prompt, allowing users to generate RGBD images from text prompts. The LDM3D model is fine-tuned on a dataset of tuples containing an RGB image, depth map and caption, and validated through extensive experiments. We also develop an application called DepthFusion, which uses the generated RGB images and depth maps to create immersive and interactive 360{\textdegree}-view experiences using TouchDesigner. This technology has the potential to transform a wide range of industries, from entertainment and gaming to architecture and design. Overall, this paper presents a significant contribution to the field of generative AI and computer vision, and showcases the potential of LDM3D and DepthFusion to revolutionize content creation and digital experiences. A short video summarizing the approach can be found at \url{https://t.ly/tdi2}.
\end{abstract}

\begin{figure*}[!t]
\centering
\includegraphics[clip, trim=0.8cm 0.5cm 0.6cm 0.4cm,width=\textwidth]{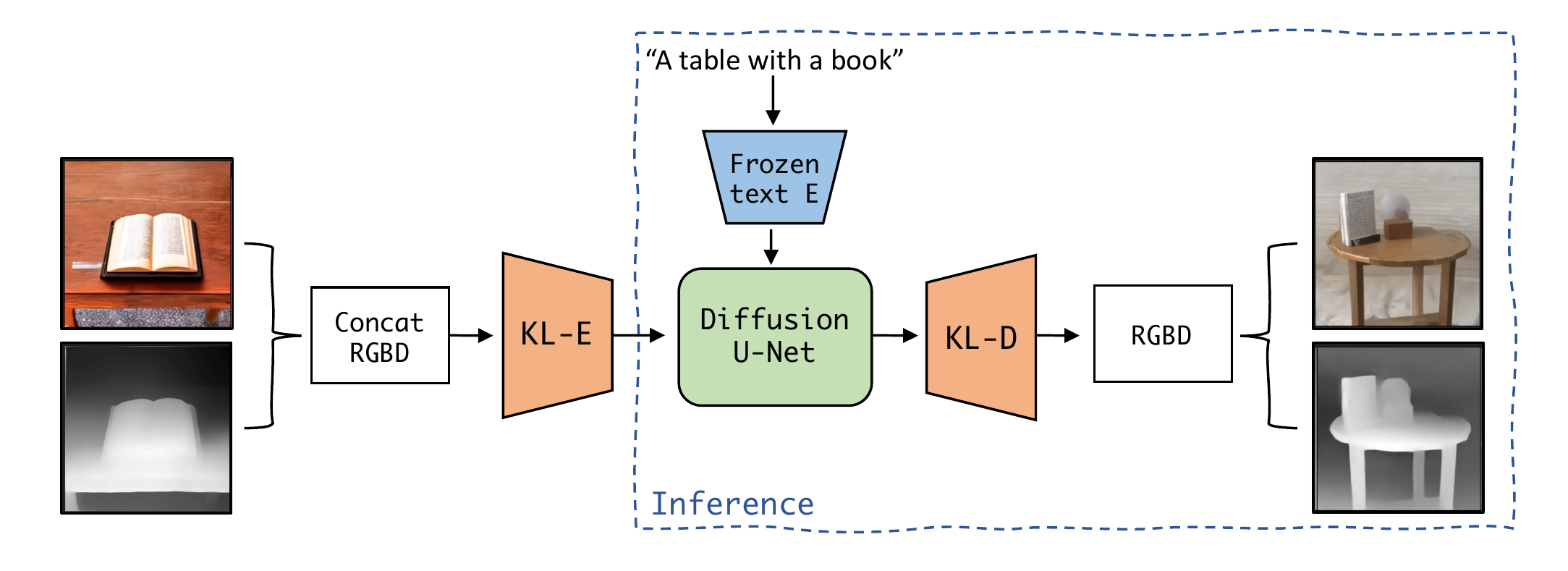} 
\caption{LDM3D overview. Illustrating the training pipeline: the 16-bit grayscale depth maps are packed into 3-channel RGB-like depth images, which are then concatenated with the RGB images along the channel dimension. This concatenated RGBD input is passed through the modified KL-AE and mapped to the latent space. Noise is added to the latent representation, which is then iteratively denoised by the U-Net model. The text prompt is encoded using a frozen CLIP-text encoder and mapped to various layers of the U-Net using cross-attention. The denoised output from the latent space is fed into the KL-decoder and mapped back to pixel space as a 6-channel RGBD output. Finally, the output is separated into an RGB image and a 16-bit grayscale depth map. Blue frame: text-to-image inference pipeline. Initiating from a Gaussian distributed noise sample in the 64x64x4-dimensional latent space. Given a text prompt, this pipeline generates an RGB image and its corresponding depth map.}
\label{fig:model_overview}
\end{figure*}

\section{Introduction}
\label{sec:intro}
The field of computer vision has seen significant advancements in recent years, particularly in the area of generative AI. In the domain of image generation, Stable Diffusion has revolutionized content creation by providing open software to generate arbitrary high-fidelity RGB images from text prompts. This work builds on top of Stable Diffusion \cite{Rombach2022latentdiffusion} v1.4 and proposes a Latent Diffusion Model for 3D (LDM3D). Unlike the original model, LDM3D is capable of generating both image and depth map data from a given text prompt as can be seen in Figure \ref{fig:model_overview}. It allows users to generate complete RGBD representations of text prompts, bringing them to life in vivid and immersive 360{\textdegree} views. 

Our LDM3D model was fine-tuned on a dataset of tuples containing an RGB image, depth map and caption. This dataset was constructed from a subset of the LAION-400M dataset, a large-scale image-caption dataset that contains over 400 million image-caption pairs. The depth maps used in fine-tuning were generated by the DPT-Large depth estimation model \cite{ranftl2020towards, Ranftl2021}, which provides highly accurate relative depth estimates for each pixel in an image. The use of accurate depth maps was crucial in ensuring that we are able to generate 360{\textdegree} views that are realistic and immersive, allowing users to experience their text prompts in vivid detail. 

To showcase the potential of LDM3D, we have developed DepthFusion, an application that uses the generated 2D RGB images and depth maps to compute a 360{\textdegree} projection using TouchDesigner \cite{TouchDesigner}. TouchDesigner is a versatile platform that enables the creation of immersive and interactive multimedia experiences. Our application harnesses the power of TouchDesigner to create unique and engaging 360{\textdegree} views that bring text prompts to life in vivid detail. DepthFusion has the potential to revolutionize the way we experience digital content. Whether it's a description of a tranquil forest, a bustling cityscape, or a futuristic sci-fi world, DepthFusion can generate immersive and engaging 360{\textdegree} views that allow users to experience their text prompts in a way that was previously impossible. This technology has the potential to transform a wide range of industries, from entertainment and gaming to architecture and design. 

In summary, our contributions are threefold. (1) We propose LDM3D, a novel diffusion model that outputs RGBD images (RGB images with corresponding depth maps) given a text prompt. (2) We develop DepthFusion, an application to create immersive 360{\textdegree}-view experiences based on RGBD images generated with LDM3D. (3) Through extensive experiments, we validate the quality of our generated RGBD images and 360{\textdegree}-view immersive videos. 

\section{Related Work}
\label{sec:related}

\paragraph{Monocular depth estimation}\hspace{-1em} is the task of estimating depth values for each pixel of a single given RGB image. 
Recent work has shown great performance in depth estimation using deep learning models based on convolutional neural networks \cite{laina2016deeper, Roy_2016_CVPR, Kuznietsov_2017_CVPR, Xu_2017_CVPR, Xu_2018_CVPR, masoumian2023gcndepth}.
Later, attention-based Transformer models were adopted to overcome the issue of a limited receptive field in CNNs, allowing the model to consider global contexts when predicting depth values \cite{ranftl2020towards, Bhat_2021_CVPR, Yang_2021_ICCV, cheng2021swin}.
Most recently diffusion models have also been applied to depth estimation to leverage the revolutionary generation capabilities of such methods.

\paragraph{Diffusion models}\hspace{-1em} have demonstrated amazing capabilities in generating highly detailed images based on an input prompt or condition \cite{pmlr-v162-nichol22a, Rombach2022latentdiffusion, saharia2022photorealistic}.
The use of depth estimates has previously been used in diffusion models as an additional condition to perform depth-to-image generation \cite{zhang2023adding}.
Later, \cite{saxena2023monocular} and \cite{duan2023diffusiondepth} showed that monocular depth estimation can also be modeled as a denoising diffusion process through the use of images as an input condition.
In this work we propose a diffusion model that simultaneously generates an RGB image and its corresponding depth map given a text prompt as input.
While our proposed model may be functionally comparable to an image generation and depth estimation model in cascade, there are several differences, challenges, and benefits of our proposed combined model.
An adequate monocular depth estimation model requires large and diverse data \cite{ranftl2020towards, MING202114}, however, as there is no depth ground truth available for generated images it is hard for off-the-shelf depth estimation models to adapt to the outputs of the diffusion model. 
Through joint training, the generation of depth is much more infused with the image generation process allowing the diffusion model to generate more detailed and accurate depth values. 
Our proposed model also differs from the standard monocular depth estimation task as the reference images are now novel images that are also generated by the model.
A similar task of generating multiple images simultaneously can be linked to video generation using diffusion models \cite{ho2022video, singer2023makeavideo, ho2022imagen}.  
Video diffusion models mostly build on \cite{ho2022video} which proposed a 3D U-Net to jointly model a fixed number of continuous frame images which are then used to compose a video. 
However, since we only require two outputs (depth and RGB) which do not necessarily require the same spatial and temporal dependencies as videos, we utilize a different approach in our model.


\section{Methodology}
\label{sec:method}
This section describes the LDM3D model's methodology, training process, and distinct characteristics that facilitate concurrent RGB image and depth map creation, as well as immersive 360-degree view generation based on LDM3D output.
\subsection{LDM-3D}


\subsubsection{Model Architecture}
LDM3D is a 1.6 billion parameter KL-regularized diffusion model, adapted from Stable Diffusion \cite{Rombach2022latentdiffusion} with minor modifications, allowing it to generate images and depth maps simultaneously from a text prompt, see \cref{fig:model_overview}.

The KL-autoencoder used in our model is a variational autoencoder (VAE) architecture based on \cite{esser2012taming}, which incorporates a KL divergence loss term. To adapt this model for our specific needs, we modified the first and last Conv2d layers of the KL-autoencoder. These adjustments allowed the model to accommodate the modified input format, which consists of concatenated RGB images and depth maps.

The generative diffusion model utilizes a U-Net backbone \cite{ronneberger2015unet} architecture, primarily composed of 2D convolutional layers. The diffusion model was trained on the learned, low-dimensional, KL-regularized latent space, similar to \cite{Rombach2022latentdiffusion}. Enabling more accurate reconstructions and efficient high-resolution synthesis, compared to transformer-based diffusion model trained in pixel space.

For text conditioning, a frozen CLIP-text encoder \cite{radford2021learning} is employed, and the encoded text prompts are mapped to various layers of the U-Net using cross-attention. This approach effectively generalizes to intricate natural language text prompts, generating high-quality images and depth maps in a single pass, only having 9,600 additional parameters compared to the reference Stable Diffusion model.

\subsubsection{Preprocessing the data} 
The model was fine-tuned on a subset of the LAION-400M \cite{laion400M} dataset, which contains image and caption pairs. The depth maps utilized in fine-tuning the LDM3D model were generated by the DPT-Large depth estimation model running inference at its native resolution of 384 $\times$ 384. Depth maps were saved in 16-bit integer format and were converted into 3-channel RGB-like arrays to more closely match the input requirements of the stable diffusion model which was pre-trained on RGB images. To achieve this conversion, the 16-bit depth data was unpacked into three separate 8-bit channels. It should be noted that one of these channels is zero for the 16-bit depth data, but this structure is designed to be compatible with a potential 24-bit depth map input. This reparametrization allowed us to encode depth information in an RGB-like image format while preserving complete depth range information.

The original RGB images and the generated RGB-like depth maps were then normalized to have values within the [0, 1] range. To create an input suitable for the autoencoder model training, the RGB images and RGB-like depth maps were concatenated along the channel dimension. This process resulted in an input image of size 512x512x6, where the first three channels correspond to the RGB image and the latter three channels represent the RGB-like depth map. The concatenated input allowed the LDM3D model to learn the joint representation of both RGB images and depth maps, enhancing its ability to generate coherent RGBD outputs.

\subsubsection{Fine-tuning Procedure.} The fine-tuning process comprises two stages, similar to the technique presented in \cite{Rombach2022latentdiffusion}. In the first stage, we train an autoencoder to generate a lower-dimensional, perceptually equivalent data representation. Subsequently, we fine-tune the diffusion model using the frozen autoencoder, which simplifies training and increases efficiency. This method outperforms transformer-based approaches by effectively scaling to higher-dimensional data, resulting in more accurate reconstructions and efficient high-resolution image and depth synthesis without the complexities of balancing reconstruction and generative capabilities.

\paragraph{Autoencoder fine-tuning.} The KL-autoencoder was fine-tuned on a training set consisting of 8233 samples, an validation set containing 2059 samples. Each sample in these sets included a caption as well as a corresponding image and depth map pair, as previously described in the preprocessing section.

For the fine-tuning of our modified autoencoder, we used a KL-autoencoder architecture with a downsampling factor of 8 time the pixel space image resolution. This downsampling factor was found to be optimal in terms of fast training process and high-quality image synthesis \cite{Rombach2022latentdiffusion}. 

During the fine-tuning process, we used the Adam optimizer with a learning rate of $10^{-5}$ and a batch size of 8. We trained the model for 83 epochs, and we sampled the outputs after each epoch to monitor the progress. The loss function for both the images and depth data consisted of a combination of perceptual loss \cite{zhang2018unreasonable} and patch-based adversarial-type loss \cite{isola2018imagetoimage}, which were originally used in the pre-training of the KL-AE \cite{esser2012taming}.

\begin{equation}
\begin{aligned}
   L_{\text{Autoencoder}} &= \min_{E, D} \max_{\psi} \bigg( L_{\text{rec}}(x, D(E(x)))\\
   &- L_{\text{adv}}(D(E(x))) + \log D_{\psi}(x)\\
   &+ L_{\text{reg}}(x; E, D) \bigg)   
\end{aligned}
\end{equation}

Here $D(E(x))$ are the reconstructed images, $L_{\text{rec}}(x, D(E(x)))$ is the perceptual reconstruction loss,  $L_{\text{adv}}(D(E(x)))$ is the adversarial loss, $D_{\psi}(x)$ is a patch based discriminator loss, and $L_{\text{reg}}(x; E, D)$ is the KL-regularisation loss.

\paragraph{Diffusion model fine-tuning}
Following the autoencoder fine-tuning, we proceeded to the second stage, which involved fine-tuning the diffusion model. This was achieved using the frozen autoencoder's latent representations as input, with a latent input size of 64x64x4. 

For this stage, we employed the Adam optimizer with a learning rate of $10^{-5}$ and a batch size of 32 . We train the diffusion model for 178 epochs with the loss function:
\begin{equation}
 L_{\text{LDM3D}} := \mathbb{E}\varepsilon(x), \epsilon \sim \mathcal{N}(0, 1),t \left[ ||\epsilon - \epsilon_\theta(z_t, t)||^2_2 \right]
\end{equation}

where $\epsilon_\theta(z_t, t)$ is the predicted noise by the denoising U-Net, and t is uniformly sampled.

We initiate the LDM3D fine-tuning using the weights from the Stable Diffusion v1.4 \cite{Rombach2022latentdiffusion} model  as a starting point. We monitor the progress throughout fine-tuning by sampling the generated images and depth maps, assessing their quality and ensuring the model's convergence.

\paragraph{Compute Infrastructure}
All training runs reported in this work are conducted on an Intel AI supercomputing cluster comprising of Intel Xeon processors and Intel Habana Gaudi AI accelerators. The LDM3D model training run is scaled out to 16 accelerators (Gaudis) on the corpus of 9,600 tupples (text caption, RGB image, depth map). The KL-autoencoder used in our LDM3D model was trained on Nvidia A6000 GPUs.

\subsubsection{Evaluation} 
In line with previous studies, we assess text-to-image generation performance using the MS-COCO \cite{lin2015microsoft} validation set. To measure the quality of the generated images, we employ Fréchet Inception Distance (FID), Inception Score (IS), and CLIP similarity metrics. The autoencoder's performance is evaluated using the relative FID score, a popular metric for comparing the quality of reconstructed images with their corresponding original input images. The evaluation was carried out on 27,265 samples, 512x512-sized from the LAION-400M dataset.
 
\begin{figure*}[!htb]
    \centering
    \begin{subfigure}{\textwidth}
        \includegraphics[width=\textwidth]{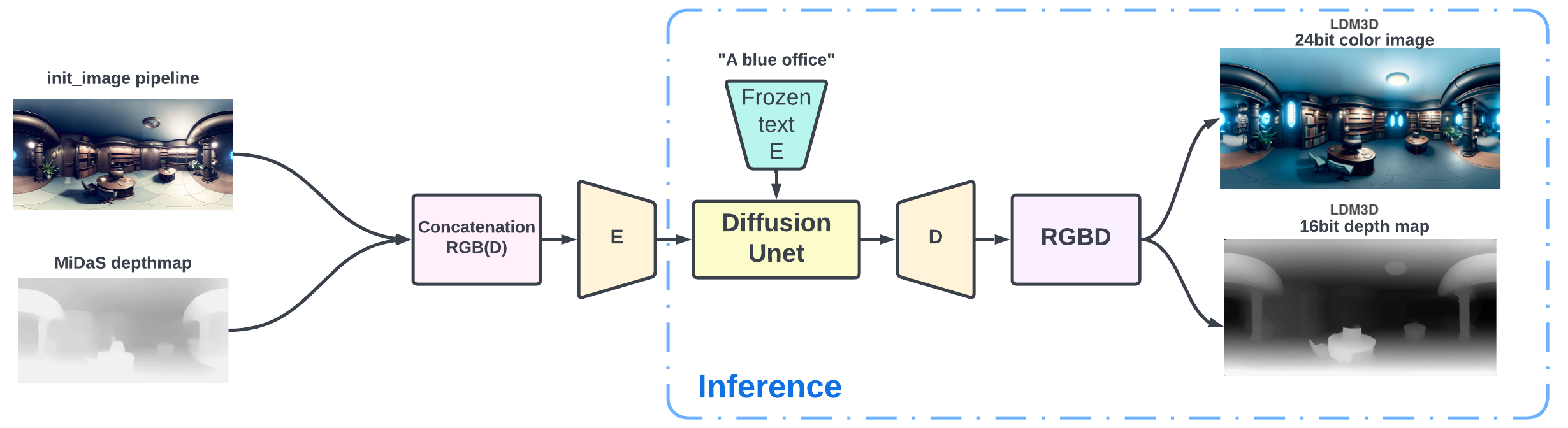}
        \caption{Step 1: Img-to-img inference pipeline for LDM3D. initiating from a panoramic image and corresponding depth map computed using DPT-Large \cite{ranftl2020towards, Ranftl2021}. The RGBD input is processed through the LDM3D image-to-image pipeline, generating a transformed image and depth map guided by the given text prompt.}
        \label{fig:360-pipeline-a}
    \end{subfigure}
    \hfill
    \vspace{12pt}
    \begin{subfigure}{\textwidth}
        \includegraphics[width=\textwidth]{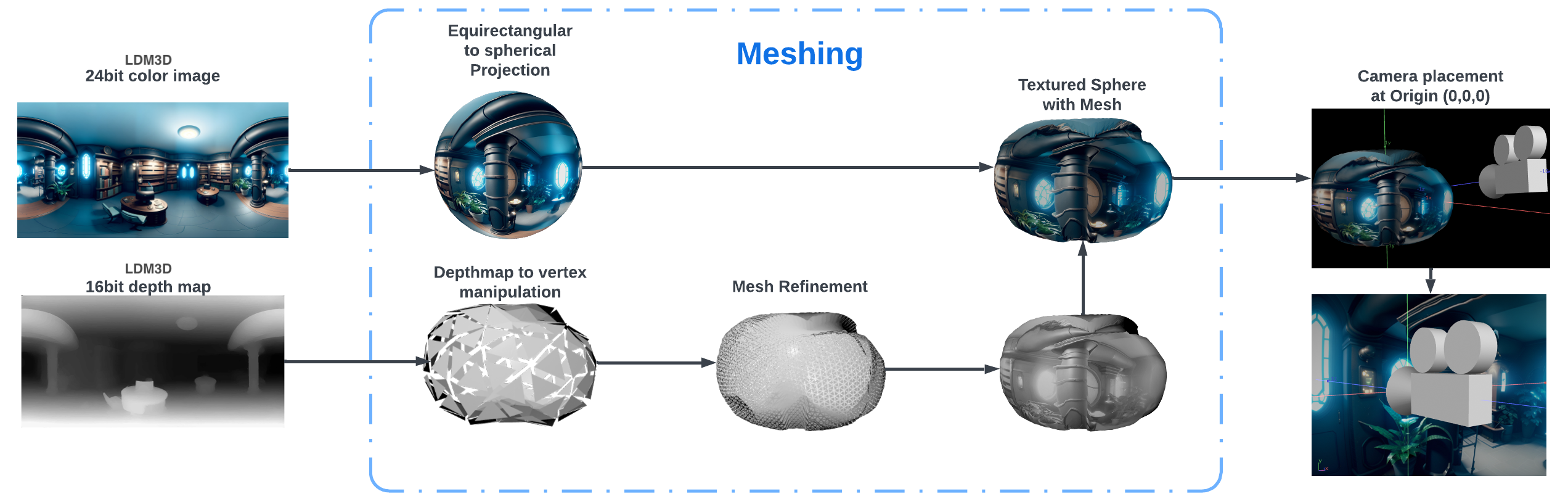}
        \caption{Step 2: LDM3D generated image is projected on a sphere, using vertex manipulation based on diffused depth map, followed by meshing.}
        \label{fig:360-pipeline-b}
    \end{subfigure}
    \hfill
    \vspace{12pt}
    \begin{subfigure}{\textwidth}
        \includegraphics[width=\textwidth]{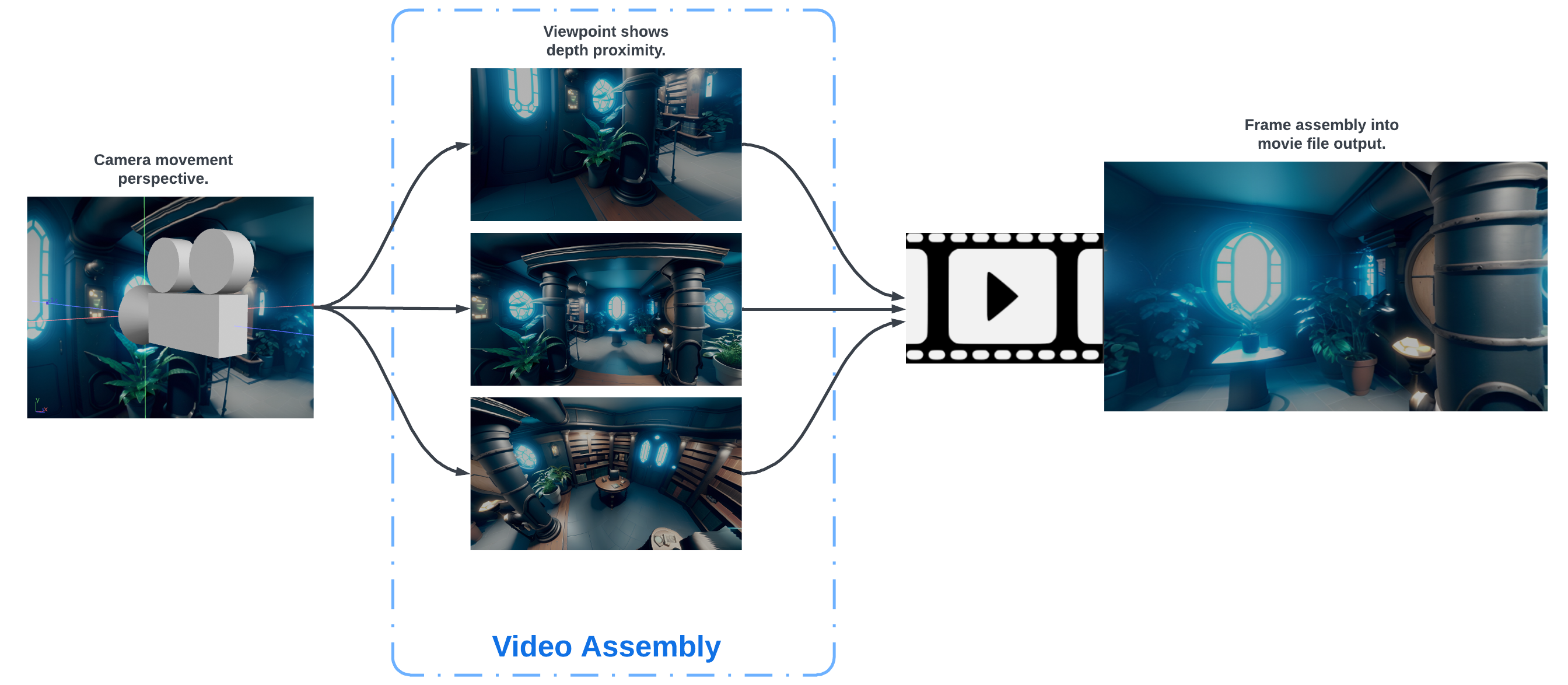}
        \caption{Step 3: Image generation from different viewpoints, and video assembly.}
        \label{fig:360-pipeline-c}
    \end{subfigure}
    \caption{Immersive experience generation pipeline.}
    \label{fig:360-pipeline}
\end{figure*}

\subsection{Immersive Experience Generation}
AI models for image generation have become prominent in the space of AI art, they are typically designed for 2D representations of diffused content. In order to project imagery onto a 3D immersive environment, modifications in mapping and resolution needed to be considered to achieve an acceptable result. Another previous limitation of correctly projected outputs occurs when perception is lost due to the monoscopic perspective of a single point of view. Modern viewing devices and techniques require disparity between two view points to achieve the experience of stereoscopic immersion. Recording devices typically capture footage from two cameras at a fixed distance so that a 3D output can be generated based on the disparity and camera parameters. In order to achieve the same from single images, however, an offset in pixel space must be calculated. With the LDM3D model, a depth map is extracted separately from RGB color space and can be used to differentiate a proper “left” and “right” perspective of the same image space in 3D. 

First, the initial image is generated and its corresponding depth map is stored, see \cref{fig:360-pipeline-a}. Using TouchDesigner  \cite{TouchDesigner}, the RGB color image is projected to the outside of an equirectangular spherical polar object in 3D space see \cref{fig:360-pipeline-b}. The perspective is set at origin 0,0,0 inside of the spherical object as the center of viewing the immersive space. The vertex points of the sphere are defined as an equal distance in all directions from the point of origin. The depth map is then used as instructions to manipulate the distance from origin to the corresponding vertex point based on monotone color values. Values closer to 1.0 move the vertex points closer to the origin, while values of 0.0 are scaled to a further distance from the origin. Values of 0.5 result in no vertex manipulation. From a monoscopic view at 0,0,0, no alteration in image can be perceived since the “rays” extend linearly from the origin outward. However, with the dual perspective of stereoscopic viewpoints, the pixels of the mapped RGB image are distorted in a dynamic fashion to give the illusion of depth. This same effect can also be observed while moving the single viewpoint away from origin 0,0,0 as the vertex distances scale equally against their initial calculation. Since the RGB color space and depth map pixels occupy the same regions, objects that have perceived geometric shapes are given approximate depth via their own virtual geometric dimensions in the render engine within TouchDesigner. \cref{fig:360-pipeline} explains the entire pipeline. This approach is not limited to the TouchDesigner platform and may also be replicated inside similar rendering engines and software suites that have the ability to utilize RGB and depth color space in their pipelines.

\begin{figure}[!htb]
  \renewcommand{\customwidth}{0.115\textwidth}
  \setlength{\tabcolsep}{1pt}
  \centering
  \begin{tabular}{c c c c}
    \multicolumn{2}{c}{\textbf{RGB Images}}  & \multicolumn{2}{c}{\textbf{Depth Maps}}  \\
    SDv1.4 & \multicolumn{2}{c}{LDM3D (Ours)} & DPT-Large \\
    \toprule
    \includegraphics[width=\customwidth]{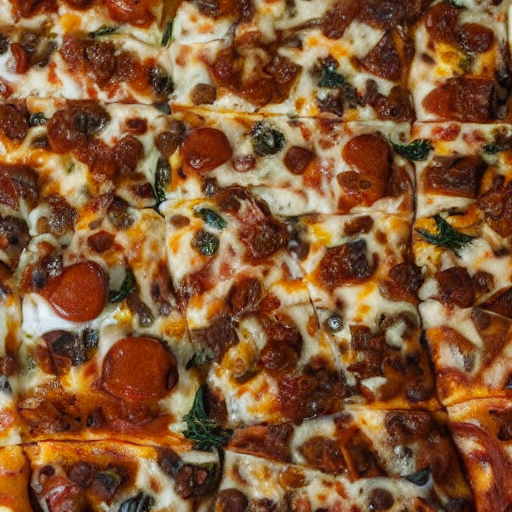} &	    \includegraphics[width=\customwidth]{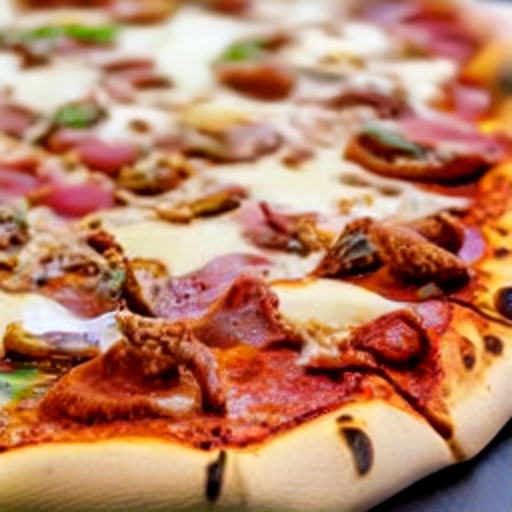} &	         
	\includegraphics[width=\customwidth]{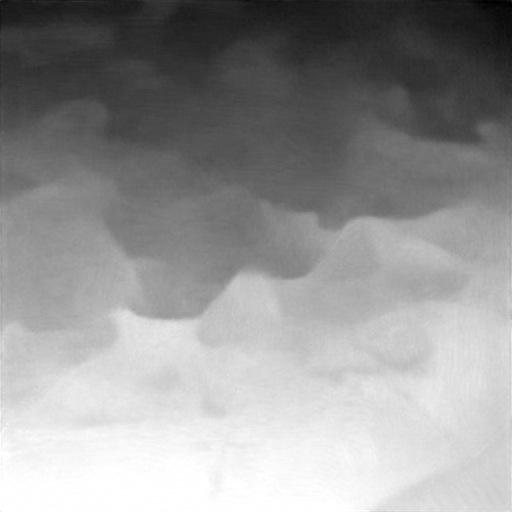} &
	\includegraphics[width=\customwidth]{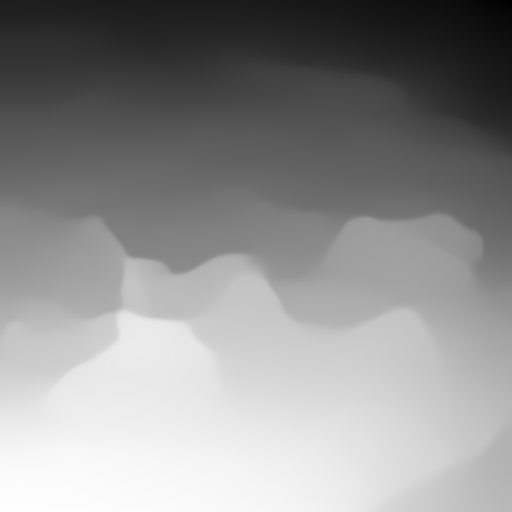} \\
 
    \includegraphics[width=\customwidth]{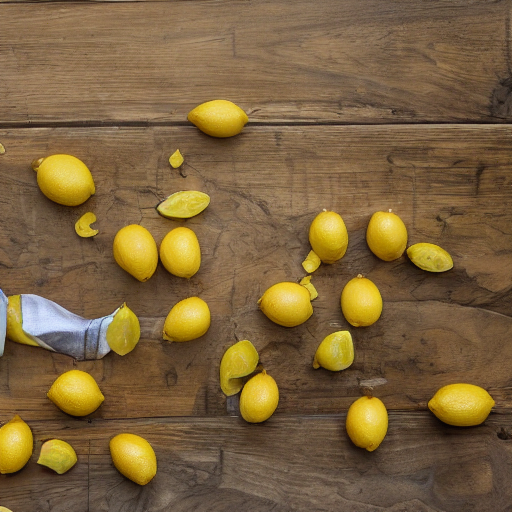} &	      \includegraphics[width=\customwidth]{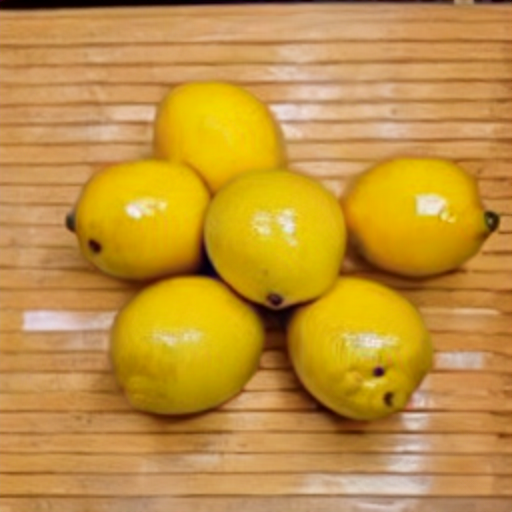} &	       
	\includegraphics[width=\customwidth]{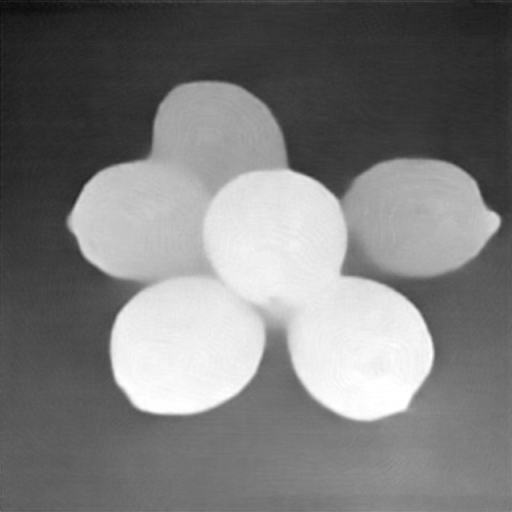} &
	\includegraphics[width=\customwidth]{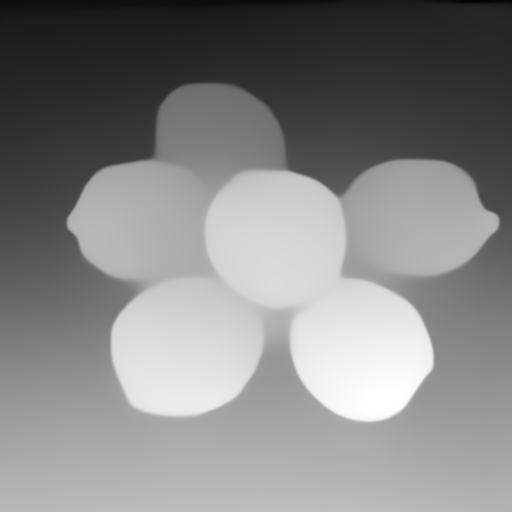} \\
 
    \includegraphics[width=\customwidth]{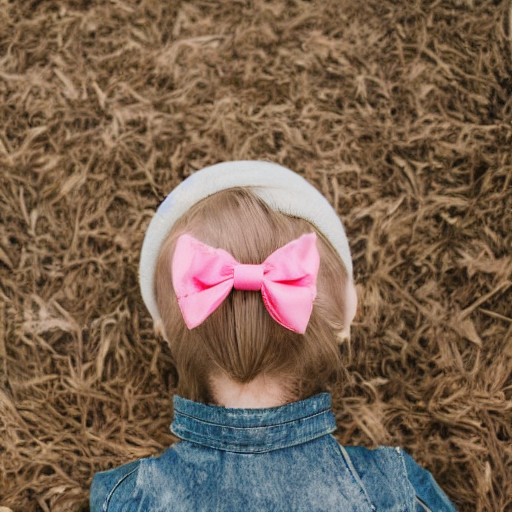} &	     
	\includegraphics[width=\customwidth]{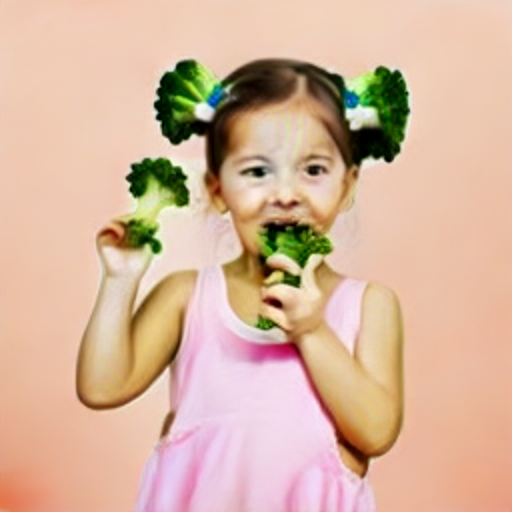} &	         
	\includegraphics[width=\customwidth]{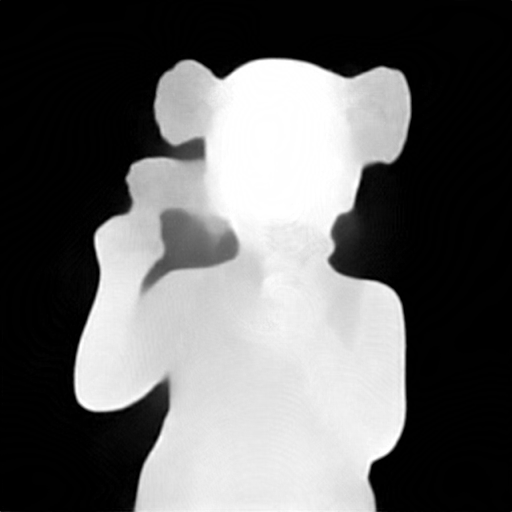} &
	\includegraphics[width=\customwidth]{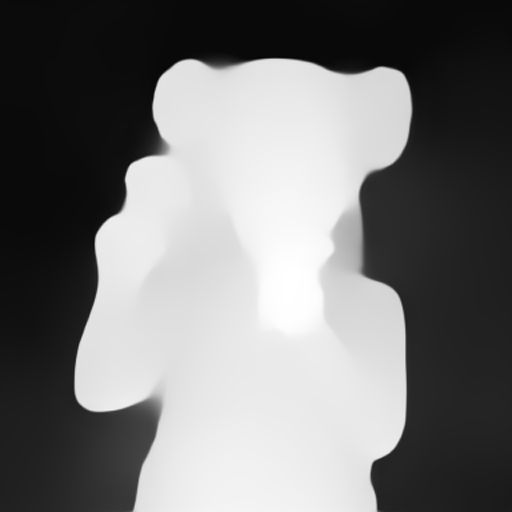} \\
 
    \includegraphics[width=\customwidth]{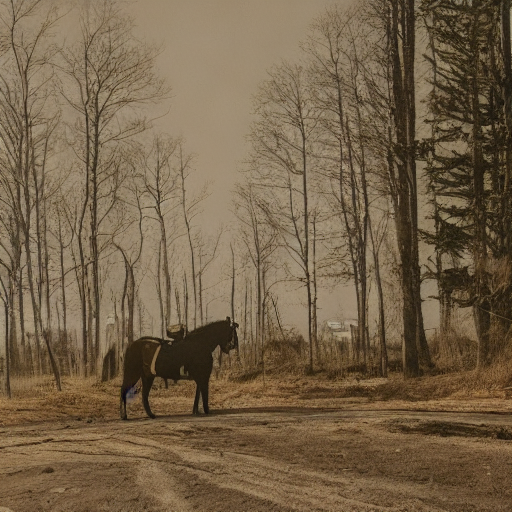} &	     
	\includegraphics[width=\customwidth]{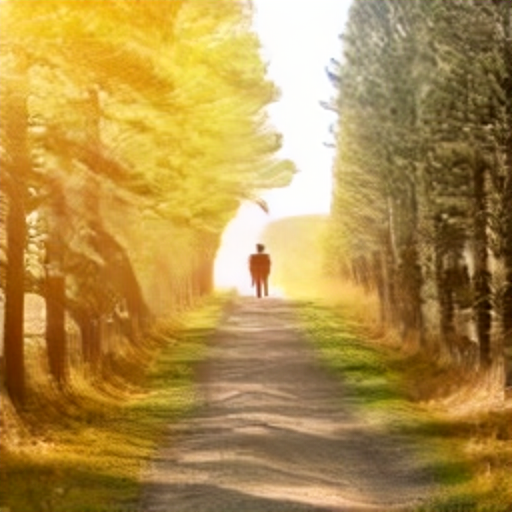} &	         
	\includegraphics[width=\customwidth]{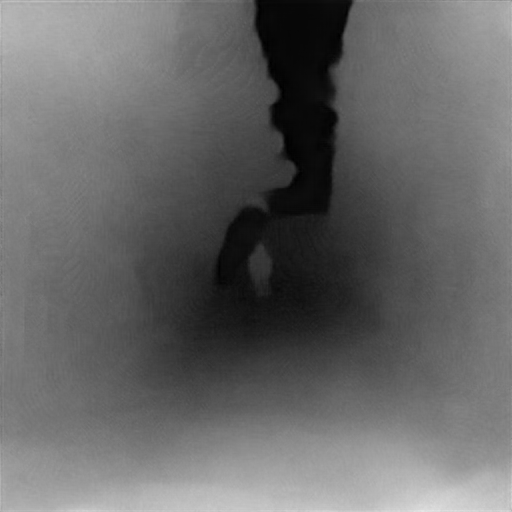} &
	\includegraphics[width=\customwidth]{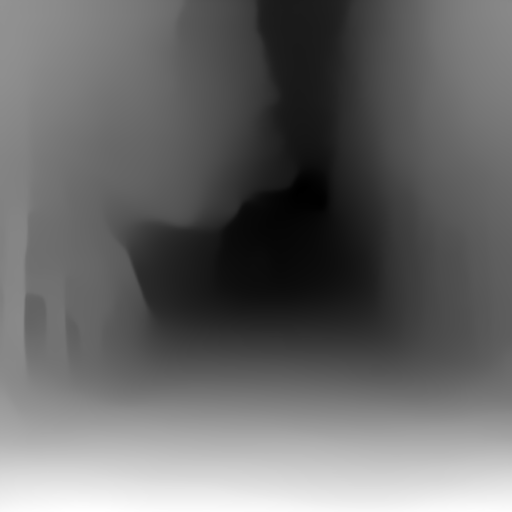} \\
 
    \includegraphics[width=\customwidth]{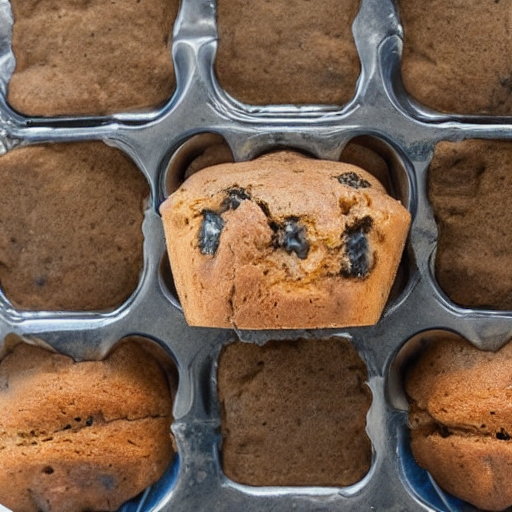} &	     
	\includegraphics[width=\customwidth]{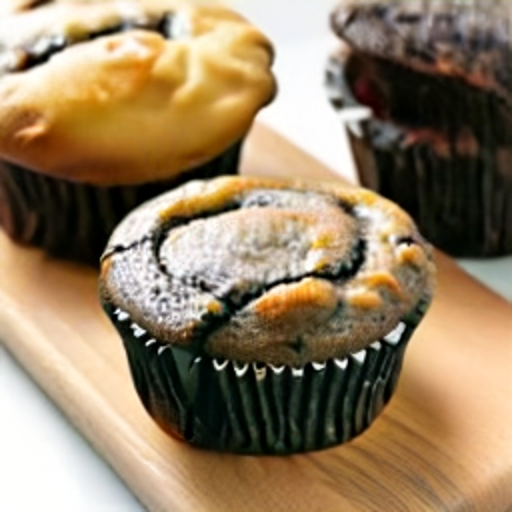} &	         
	\includegraphics[width=\customwidth]{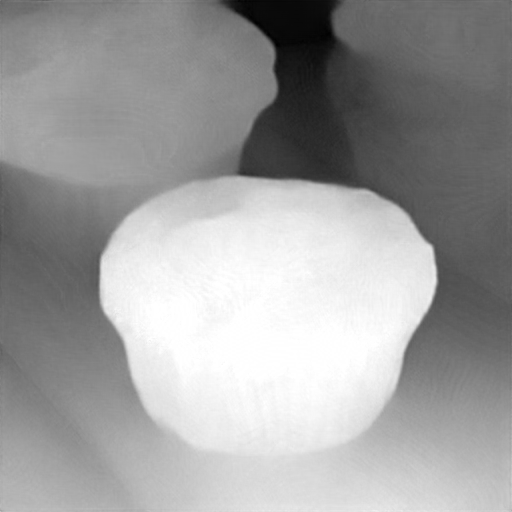} &
	\includegraphics[width=\customwidth]{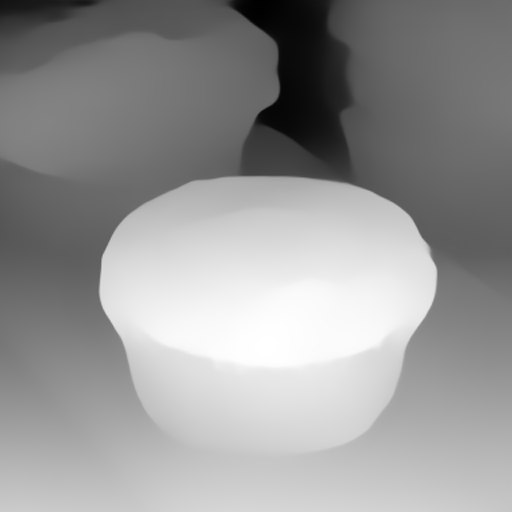} \\
  
    \includegraphics[width=\customwidth]{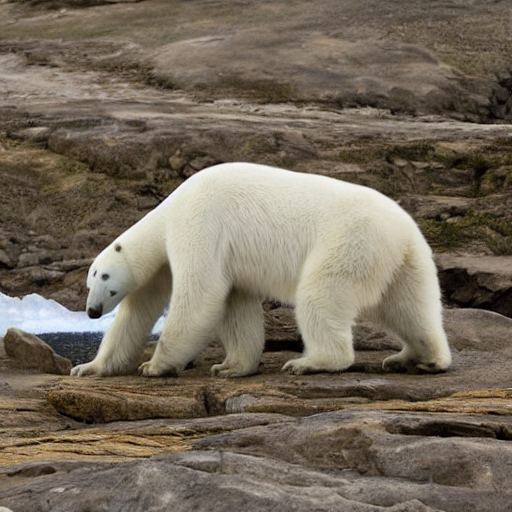} &	         
	\includegraphics[width=\customwidth]{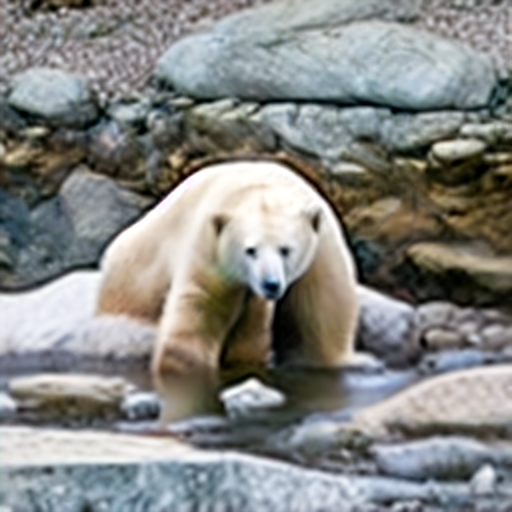} &	         
	\includegraphics[width=\customwidth]{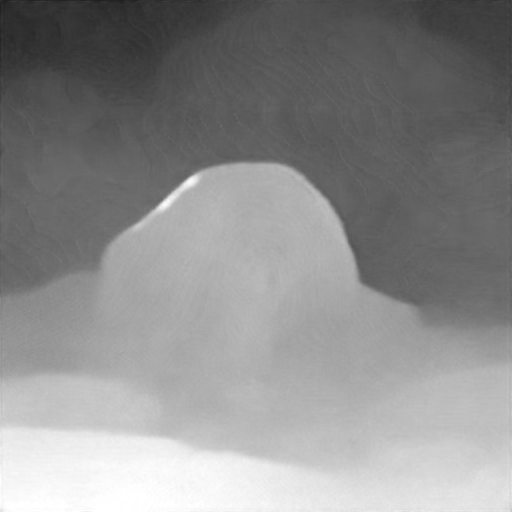} &
	\includegraphics[width=\customwidth]{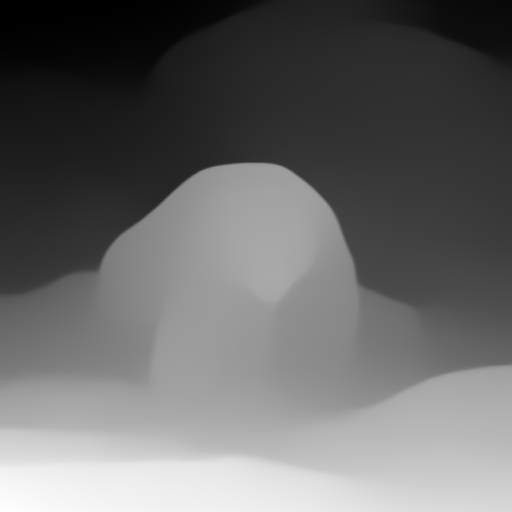}
    \\
    \bottomrule  
  \end{tabular}    
  \caption{Qualitative comparison of images to Stable diffusion v1.4 \cite{Rombach2022latentdiffusion} and depth maps to DPT-Large \cite{ranftl2020towards, Ranftl2021}, on $512 \times 512$ images from the COCO validation dataset. Captions from top to bottom:"a close up of a sheet of pizza on a table", "A picture of some lemons on a table", "A little girl with a pink bow in her hair eating broccoli", "A man is on a path riding a horse","A muffin in a black muffin wrap next to a fork", "a white polar bear drinking water from a water source next to some rocks".}
  \label{fig:oursvsSD}
\end{figure}

\section{Results}
\label{sec:exp}
In the following, we show the high quality of the generated images and depth maps of our LDM3D model. We also show the impact on performance of the autoencoder when adding the depth modality.


\subsection{Qualitative Evaluation} 
A qualitative analysis of the generated images and depth maps reveals that our LDM3D model can effectively generate visually coherent outputs that correspond well to the provided text prompts. The generated images exhibit fine details and complex structures, while the depth maps accurately represent the spatial information of the scenes, see \cref{fig:oursvsSD} . These results highlight the potential of our model for various applications, including 3D scene reconstruction and immersive content creation, see \cref{fig:360-pipeline}. 
A video with examples of the immersive 360-views that can be generated using our complete pipeline can be found at \url{https://t.ly/TYA5A}.

\subsection{Quantitative Image Evaluation} 
Our LDM3D model demonstrates impressive performance in generating high-quality images and depth maps from text prompts. When evaluated on the MS-COCO validation set, the model achieves competitive scores to the Stable diffusion baseline using FID and CLIP similarity metrics, see \cref{table:methods_comparison}. There is a degradation in the inception score (IS), which might indicate that our model generates images that are close to the real images in terms of their feature distributions, as could be derived by the similar FID scores, but they might lack diversity or some aspects of image quality that IS captures. Nevertheless, IS is considered to be a less robust metric than FID because it struggles with capturing intra-class diversity \cite{borji2021pros}, is highly sensitive to model parameters and implementations, whereas FID is better at assessing the similarity between distributions of real and generated images while being less sensitive to minor changes in network weights that don't impact image quality \cite{barratt2018note}. The high CLIP similarity score indicates that the model maintains a high level of detail and fidelity with respect to the text prompts.

\pgfplotsset{width=7.5cm,compat=1.3}


\begin{table}[H]
    \centering
    \begin{tabular}{lccc}
        \toprule
        Method    & FID$\downarrow$  & IS$\uparrow$  & CLIP$\uparrow$  \\
        \midrule
        SD v1.4  & 28.08 & \textbf{34.17}$\pm0.76$ & $26.13\pm2.81$       \\
        SD v1.5  & \textbf{27.39} & $34.02\pm0.79$ & $26.13\pm2.79$      \\
        \midrule
        LDM3D (ours)  & 27.82 & $28.79\pm0.49$ & \textbf{26.61}$\pm2.92$         \\
        \bottomrule
    \end{tabular}
    \caption{Text-to-Image synthesis. Evaluation of text-conditional image synthesis on the
512 x 512-sized MS-COCO \cite{lin2015microsoft} dataset with 50 DDIM \cite{song2021denoising} steps. Our model is on par with the Stable diffusion models with the same number of parameters (1.06B). IS and CLIP similarity scores are averaged over 30k captions from the MS-COCO dataset.}
    \label{table:methods_comparison}
\end{table}

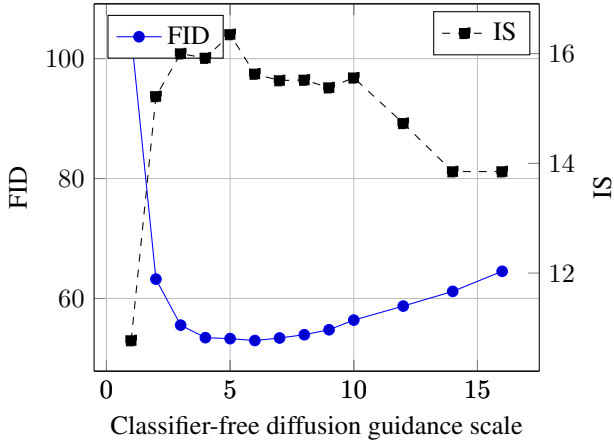
\begin{figure}[!htb]
    \centering
    \begin{tikzpicture}
        \begin{axis}[
            axis y line*=left,
            xlabel= {Classifier-free diffusion guidance scale},
            ylabel= {FID},
            grid=major,
            legend pos=north west,
            ]
            \addplot coordinates {
                (1,104.03)
                (2,63.23)
                (3,55.55)
                (4,53.46)
                (5,53.30)
                (6,52.98)
                (7,53.41)
                (8,53.95)
                (9,54.78)
                (10,56.39)
                (12,58.73)
                (14,61.18)
                (16,64.53)
            };
            \addlegendentry{FID}
            \label{plot:left_}
        \end{axis}
        \begin{axis}[
            axis y line*=right,
            xlabel= ,
            ylabel= {IS},
            ]
            \addplot[
            color=black,           
            mark=square*,         
            dashed,               
            ]
            coordinates {
                (1,10.77)
                (2,15.22)
                (3,16.0)
                (4,15.92)
                (5,16.35)
                (6,15.63)
                (7,15.51)
                (8,15.52)
                (9,15.38)
                (10,15.56)
                (12,14.73)
                (14,13.85)
                (16,13.85)
            };
            \addlegendentry{IS}
            \label{plot:right_}
        \end{axis}
    \end{tikzpicture}
    \caption{FID / IS vs. Classifier-free diffusion guidance scale factor. Evaluation of text-conditional image synthesis on 2000 samples,
512 x 512-sized from MS-COCO \cite{lin2015microsoft} dataset, with 50 DDIM \cite{song2021denoising} steps.}
    \label{fig:fid_vs_scale}
\end{figure}

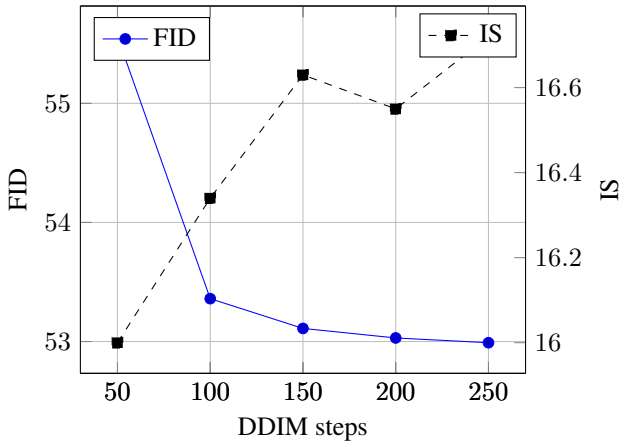
\begin{figure}[!htb]
    \centering
    \begin{tikzpicture}
        \begin{axis}[
            axis y line*=left,
            xlabel= {DDIM steps},
            ylabel= {FID},
            grid=major,
            legend pos=north west,
        ]
            \addplot coordinates {
                (50,55.56)
                (100,53.36)
                (150,53.11)
                (200,53.03)
                (250,52.99)
            };
            \addlegendentry{FID}
            \label{plot:left}
        \end{axis}
        \begin{axis}[
            axis y line*=right,
            xlabel= ,
            ylabel= {IS},
        ]
            \addplot [
            color=black,           
            mark=square*,         
            dashed,               
            ] 
            coordinates {
                (50,16.0)
                (100,16.34)
                (150,16.63)
                (200,16.55)
                (250,16.72)
            };
            \addlegendentry{IS}
            \label{plot:left__}
        \end{axis}
    \end{tikzpicture}
    \caption{FID / IS vs. DDIM steps. Evaluation of text-conditional image synthesis on 2000 samples,
512 x 512-sized from MS-COCO \cite{lin2015microsoft} dataset, s=3.}
    \label{fig:fid_vs_ddim_step}
\end{figure}

\begin{figure}[!htb]
    \centering
    \begin{tikzpicture}
        \begin{axis}[
            xlabel=Training Step,
            ylabel=FID,
            grid=major,
        ]
            \addplot coordinates {
                (2056,72.81)				
                (14135,67.53)
                (26214,61.58)
                (40092,58.87)
                (46003,58.53)
            };
        \end{axis}
    \end{tikzpicture}
    \caption{FID vs. Training Step. Evaluation of text-conditional image synthesis on 2000 samples,
512 x 512-sized from MS-COCO \cite{lin2015microsoft} dataset: with 50 DDIM \cite{song2021denoising} steps, s=3.}
    \label{fig:fid_vs_training_step}
\end{figure}
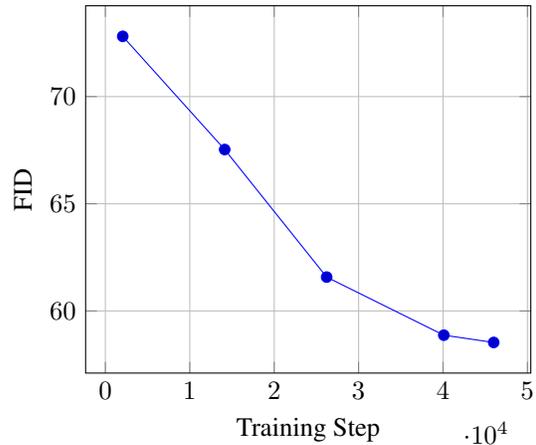

In addition, we investigate the relationship between key hyperparameters and the quality of the generated images. We plot the FID and IS scores against the classifier-free diffusion guidance scale factor (\cref{fig:fid_vs_scale}), the number of denoising steps (\cref{fig:fid_vs_ddim_step}), and the training step (\cref{fig:fid_vs_training_step}). Additionally, we plotted the CLIP similarity score against the classifier-free diffusion guidance scale factor in see \cref{fig:clip_sim_vs_scale}.

\cref{fig:fid_vs_scale} indicates that the optimal classifier-free diffusion guidance scale factor that produces the best balance between image quality and diversity is around s=5, higher than reported on Stable diffusion v1.4 (s=3). \cref{fig:clip_sim_vs_scale} indicated that the alignment of the generated images with the input text prompts is nearly unaffected as the scale factor changes for scale factors larger than 5.

\cref{fig:fid_vs_ddim_step} indicates that the image quality increases with the number of denoising steps, the most significant improvement occurs when increasing the DDIM steps from 50 to 100.


\begin{figure}[!htb]
    \centering
    \begin{tikzpicture}
        \begin{axis}[
            xlabel=Classifier-free diffusion guidance scale,
            ylabel=CLIP similarity,
            grid=major,
        ]
            \addplot coordinates {
                (1,24.10)
                (2,26.04)
                (3,26.44)
                (4,26.58)
                (5,26.71)
                (6,26.82)
                (7,26.80)
                (8,26.84)
                (9,26.85)
                (10,26.78)
                (12,26.83)
                (14,26.80)
                (16,26.74)
            };
        \end{axis}
    \end{tikzpicture}
    \caption{CLIP similarity score vs. Classifier-free diffusion guidance scale factor. Averaged on 2000 samples,
512 x 512-sized generated from MS-COCO \cite{lin2015microsoft} dataset captions, with 50 DDIM \cite{song2021denoising} steps.}
    \label{fig:clip_sim_vs_scale}
\end{figure}
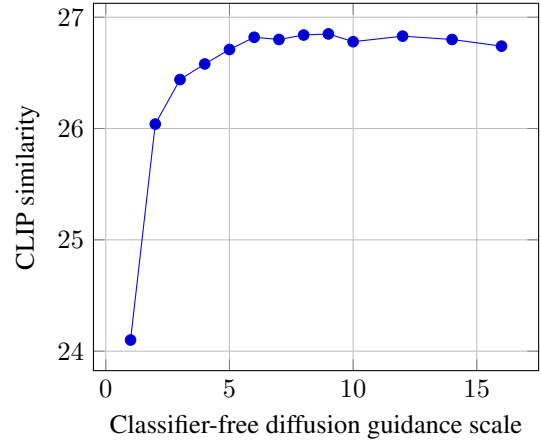

\subsection{Quantitative Depth Evaluation}

\begin{figure}[]
\centering
\includegraphics[width=\linewidth]{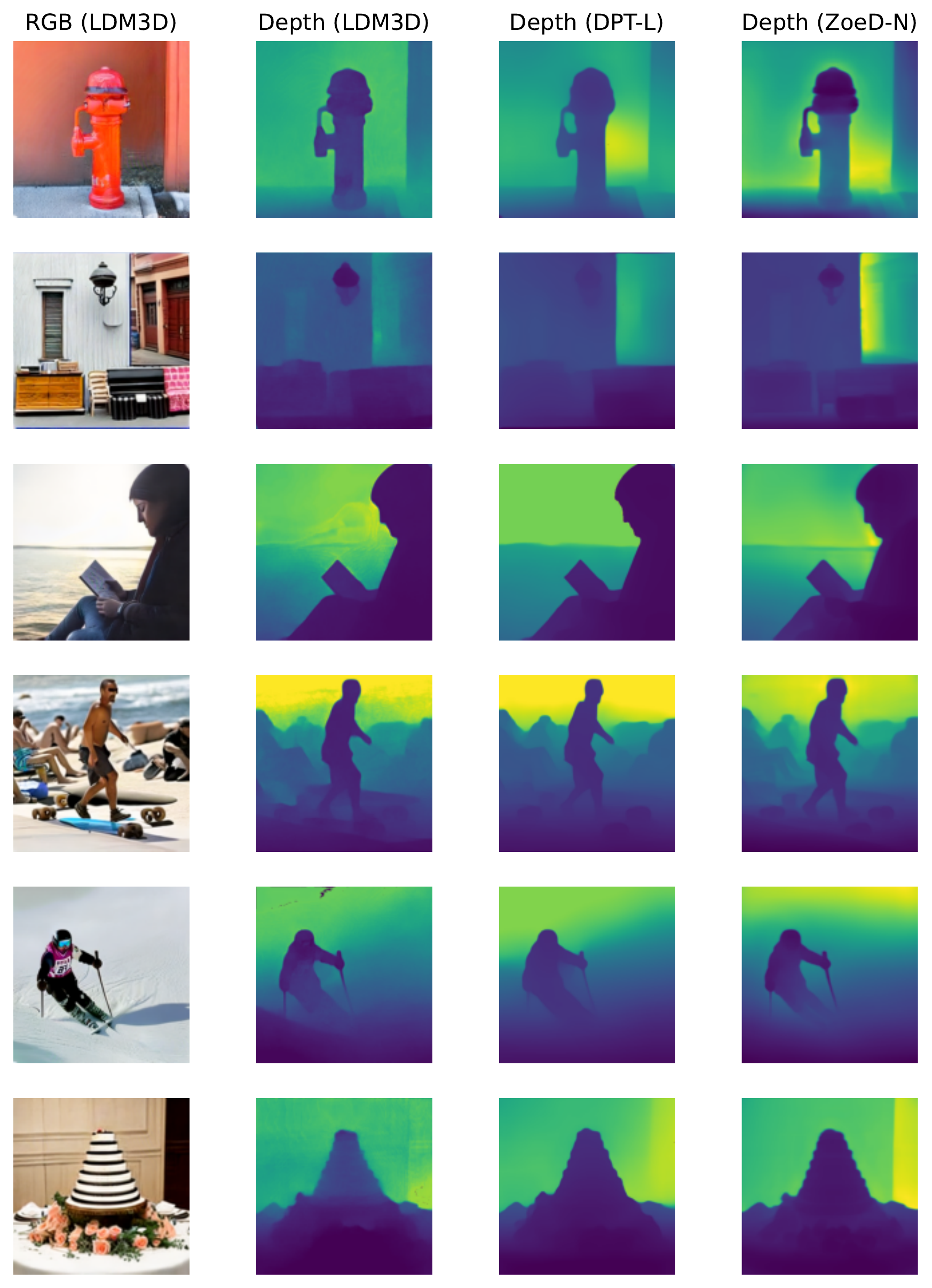} 
\caption{Depth visualization to accompany \cref{table:depth_metrics_comparison}.}
\label{fig:depth_visual_comparison}
\end{figure}

Our LDM3D model jointly outputs images and their corresponding depth maps. Since there is no ground truth depth reference for these images, we define a reference model against which to compute depth metrics. For this, we select the ZoeDepth metric depth estimation model. LDM3D outputs depth in disparity space, as it was fine-tuned using depth maps produced by DPT-Large. We align these depth maps to reference ones produced by ZoeDepth. This alignment is done in disparity space in a global least-squares fitting manner similar to the approach in~\cite{ranftl2020towards}. Points to be fitted to are determined via random sampling applied to the intersected validity maps of the estimated and target depth maps, where valid depth is simply defined to be non-negative. The alignment procedure computes per-sample scale and shift factors that are applied to the LDM3D and DPT-Large depth maps to align the depths to ZoeDepth values. All depth maps are then inverted to bring them into metric depth space. The two depth metrics we compute are absolute relative error (AbsRel) and root mean squared error (RMSE). Metrics are aggregated over a 6k subset of images from the 30k set used for image evaluation. \cref{table:depth_metrics_comparison} shows that LDM3D achieves similar depth accuracy as DPT-Large, demonstrating the success of our finetuning approach. A corresponding visualization is shown in \cref{fig:depth_visual_comparison}.

\begin{table}[]
    \centering
    \begin{tabular}{l|cc}
    \toprule
     & \multicolumn{2}{c}{\textit{w.r.t. ZoeDepth-N}} \\
     & AbsRel &  RMSE [m]\\
    \midrule
    LDM3D & 0.0911 & 0.334 \\
    DPT-Large & 0.0779 & 0.297 \\
    \bottomrule
    \end{tabular}
    \caption{Depth evaluation comparing LDM3D and DPT-Large with respect to ZoeDepth-N that serves as a reference model.}  
    \label{table:depth_metrics_comparison}
\end{table}

\begin{table}[h!]
    \centering
    \begin{tabular}{lcc}
        \toprule
        \textbf{Model} & \textbf{rFID} & \textbf{Abs.Rel.} \\
        \midrule
        pre-trained KL-AE, RGB & 0.763 & - \\
        fine-tuned KL-AE, RGBD & 1.966 & 0.179 \\
        \bottomrule
    \end{tabular}
    \caption{Comparison of KL-autoencoder fine-tuning approaches. The pre-trained KL-AE was evaluated on 31,471 images, and the fine-tuned KL-AE on 27,265 images, 512x512-sized from the LAION-400M \cite{laion400M} dataset.}
    \label{table:kl_autoencoder_fine_tuning}
\end{table}


\subsection{Autoencoder Performance}
We first evaluate the performance of our fine-tuned KL-AE using the relative FID score, see \cref{table:kl_autoencoder_fine_tuning}. Our findings show a minor but measurable decline in the quality of reconstructed images compared to the pre-trained KL-AE. This can be attributed to the increased data compression ratio when incorporating depth information alongside RGB images in the pixel space, but keeping the latent space dimensions unchanged. Note that the adjustments made to the AE are minimal, adding only 9,615 parameters to the pre-trained AE. We expect that further modifications to the AE can further improve performance. In the current architecture, this decrease in quality is compensated by  fine-tuning the diffusion U-Net. The resulting LDM3D model performs on par with vanilla Stable Diffusion as shown in the previous sections.

\section{Conclusion}
\label{sec:conclusion}
In conclusion, this research paper introduces LDM3D, a novel diffusion model that generates RGBD images from text prompts. To demonstrate the potential of LDM3D we also develop DepthFusion, an application that creates immersive and interactive 360-view experiences using the generated RGBD images in TouchDesigner. The results of this research have the potential to revolutionize the way we experience digital content, from entertainment and gaming to architecture and design. The contributions of this paper pave the way for further advancements in the field of multi-view generative AI and computer vision. We look forward to seeing how this space will continue to evolve and hope that the presented work will be useful for the community.

{\small
\bibliographystyle{ieee_fullname}
\bibliography{egbib}
}

\end{document}